\documentclass[sigconf,nonacm]{acmart}

\AtBeginDocument{%
  \providecommand\BibTeX{{%
    \normalfont B\kern-0.5em{\scshape i\kern-0.25em b}\kern-0.8em\TeX}}}

\acmConference[Preprint]{Preprint}
\usepackage{comment}
\begin{document}

\title{Knowledge-Infused Self Attention Transformers}

\author{Kaushik Roy}
\email{kaushikr@email.sc.edu}
\affiliation{%
  \institution{Artificial Intelligence Institute \\ University of South Carolina}
  \country{USA}
}
\author{Yuxin Zi}
\email{yzi@email.sc.edu}
\affiliation{%
  \institution{Artificial Intelligence Institute \\ University of South Carolina}
  \country{USA}
}
\author{Vignesh Narayanan}
\email{vignar@sc.edu}
\affiliation{%
  \institution{Artificial Intelligence Institute \\ University of South Carolina}
  \country{USA}
}
\author{Manas Gaur}
\email{manas@umbc.edu}
\affiliation{%
  \institution{KAI\textsuperscript{2}, University of Maryland \\ Baltimore County}
  \country{USA}
}
\author{Amit Sheth}
\email{amit@sc.edu}
\affiliation{%
  \institution{Artificial Intelligence Institute \\ University of South Carolina}
  \country{USA}
}

\begin{abstract}
Transformer-based language models have achieved impressive success in various natural language processing tasks due to their ability to capture complex dependencies and contextual information using self-attention mechanisms. However, they are not without limitations. These limitations include hallucinations, where they produce incorrect outputs with high confidence, and alignment issues, where they generate unhelpful and unsafe outputs for human users. These limitations stem from the absence of implicit and missing context in the data alone. To address this, researchers have explored augmenting these models with external knowledge from knowledge graphs to provide the necessary additional context. However, the ad-hoc nature of existing methods makes it difficult to properly analyze the effects of knowledge infusion on the many moving parts or components of a transformer. This paper introduces a systematic method for infusing knowledge into different components of a transformer-based model. A modular framework is proposed to identify specific components within the transformer architecture, such as the self-attention mechanism, encoder layers, or the input embedding layer, where knowledge infusion can be applied. Additionally, extensive experiments are conducted on the General Language Understanding Evaluation (GLUE) benchmark tasks, and the findings are reported. This systematic approach aims to facilitate more principled approaches to incorporating knowledge into language model architectures.
\end{abstract}





\maketitle
\section{Introduction}\label{sec:intro}
Language modeling has witnessed significant advancements with the introduction of self-attention-based transformer architectures (e.g., GPT-3, ChatGPT, PaLM, etc.)\cite{brown2020language,chowdhery2022palm}. These models have achieved remarkable success in a wide range of natural language processing tasks, demonstrating their ability to generate coherent and contextually relevant text. By utilizing self-attention mechanisms, transformers excel at capturing long-range dependencies and establishing meaningful relationships between words, enabling them to generate high-quality, context-aware text. However, despite their successes, self-attention-based transformer models have limitations when it comes to capturing all the necessary context solely from the available data. Language models often struggle with comprehending missing or implicit information, particularly in scenarios where the training data is incomplete or lack the desired context\cite{zhu2023knowledge}. This limitation can lead to generated text that is plausible but semantically incorrect or inconsistent, diminishing the model's ability to fully understand and generate language with nuanced meaning\cite{zhou2020evaluating}. To address these limitations, incorporating external knowledge into language models can provide the missing and implicit context required for accurate language generation. External knowledge, such as factual information, world knowledge, or domain-specific expertise, can supplement the training data by offering additional context that may not be explicitly present in the data alone. By integrating external knowledge, language models can enhance their understanding of complex concepts, disambiguate ambiguous statements, and generate more coherent and contextually accurate text.

However, the existing methods used to incorporate external knowledge into language models often lack a systematic and well-defined approach. These methods seem rather ad hoc, as they introduce knowledge at various components of the transformer architecture based mainly on empirical justifications related to improved performance in downstream tasks. Transformers comprise several interconnected components, including input embedding matrices, encoder layers, and self-attention operations. One concern is that augmenting knowledge in an ad hoc manner may lead to the exploitation of statistical artifacts by the numerous moving parts of the transformer\cite{mccoy2019right}. For instance, it could involve overfitting by utilizing additional parameters provided by the knowledge or fitting to task-specific hidden or spurious patterns to achieve high downstream performance scores. Consequently, it remains unclear, based solely on performance metrics, to what extent such augmentation truly enhances the language comprehension and understanding of the model\cite{bender2020climbing}.

\begin{figure}[!htb]
    \centering
    \includegraphics[width=\linewidth,trim = 1.5cm 1cm 0cm 0cm, clip]{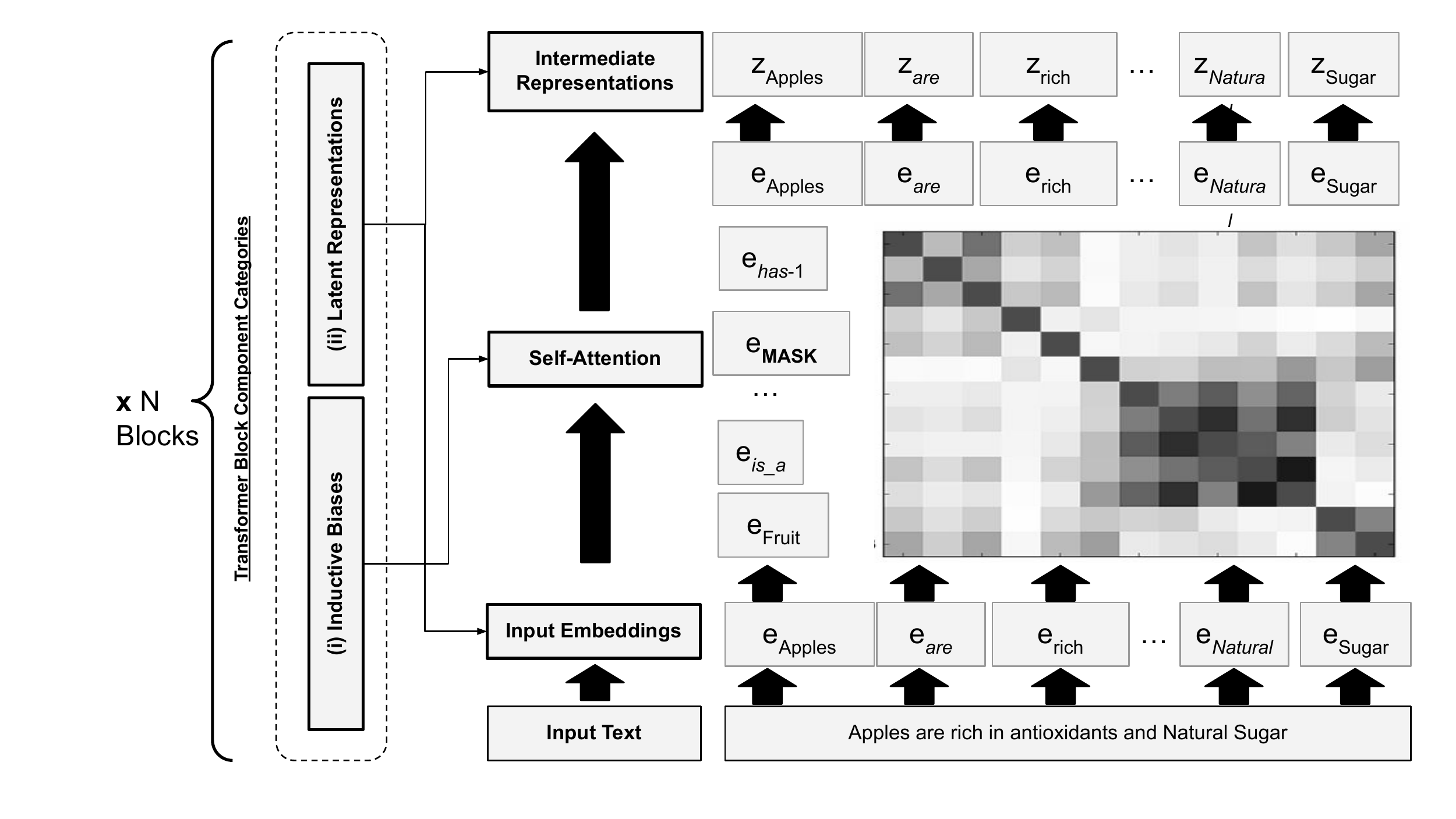}
    \caption{The figure shows the different components of a transformer block, categorized into inductive biases and latent representations. We categorize the input embeddings and the representations obtained after each encoder as latent representations and the self-attention matrices as inductive biases. This structure is repeated N times (e.g., 12 times in BERT).}
    \label{fig:transformer}
\end{figure}

In light of these limitations, we propose a systematic approach to infusing knowledge in language models that adopts different strategies depending on the specific transformer component. Initially, we categorize the architectural elements of a transformer into two groups: (i) inductive biases, such as the self-attention matrices, and (ii) latent representations, including the input embeddings and the intermediate representations between encoder layers (Figure \ref{fig:transformer} illustrates this categorization). Subsequently, we introduce three distinct categories of knowledge infusion: (i) shallow knowledge infusion, which involves incorporating knowledge at the latent representations of the first transformer block; (ii) semi-deep knowledge infusion, where knowledge is also integrated at the self-attention matrix (inductive bias) of the first transformer block, and (iii) deep knowledge infusion, which interleaves knowledge incorporation at the latent representations and self-attention matrices (inductive biases) of the various transformer blocks. In essence, these three methods aim to infuse knowledge at either the level of inductive biases, latent representations, or both. To evaluate the effectiveness of the three categories of knowledge infusion, we conduct extensive experimentation and ablation studies on various tasks from the General Language Understanding Evaluation (GLUE) benchmark, reporting our findings\cite{wang2018glue}. As mentioned earlier, due to the potential exploitation of statistical artifacts by knowledge infusion methods, traditional downstream task performance metrics can be an ineffective measure of knowledge infusion. Therefore, we also introduce new evaluation metrics for a more robust measurement of the effectiveness of knowledge infusion (see Section \ref{subsec:eval_met}). Our results indicate that deep knowledge infusion outperforms the other two categories of knowledge infusion using both traditional metrics of accuracy and F1-score, as well as the newly introduced metrics.
\section{Knowledge-Infused Self-Attention Transformers}
\begin{figure*}[!htb]
    \centering
    \includegraphics[width=\linewidth]{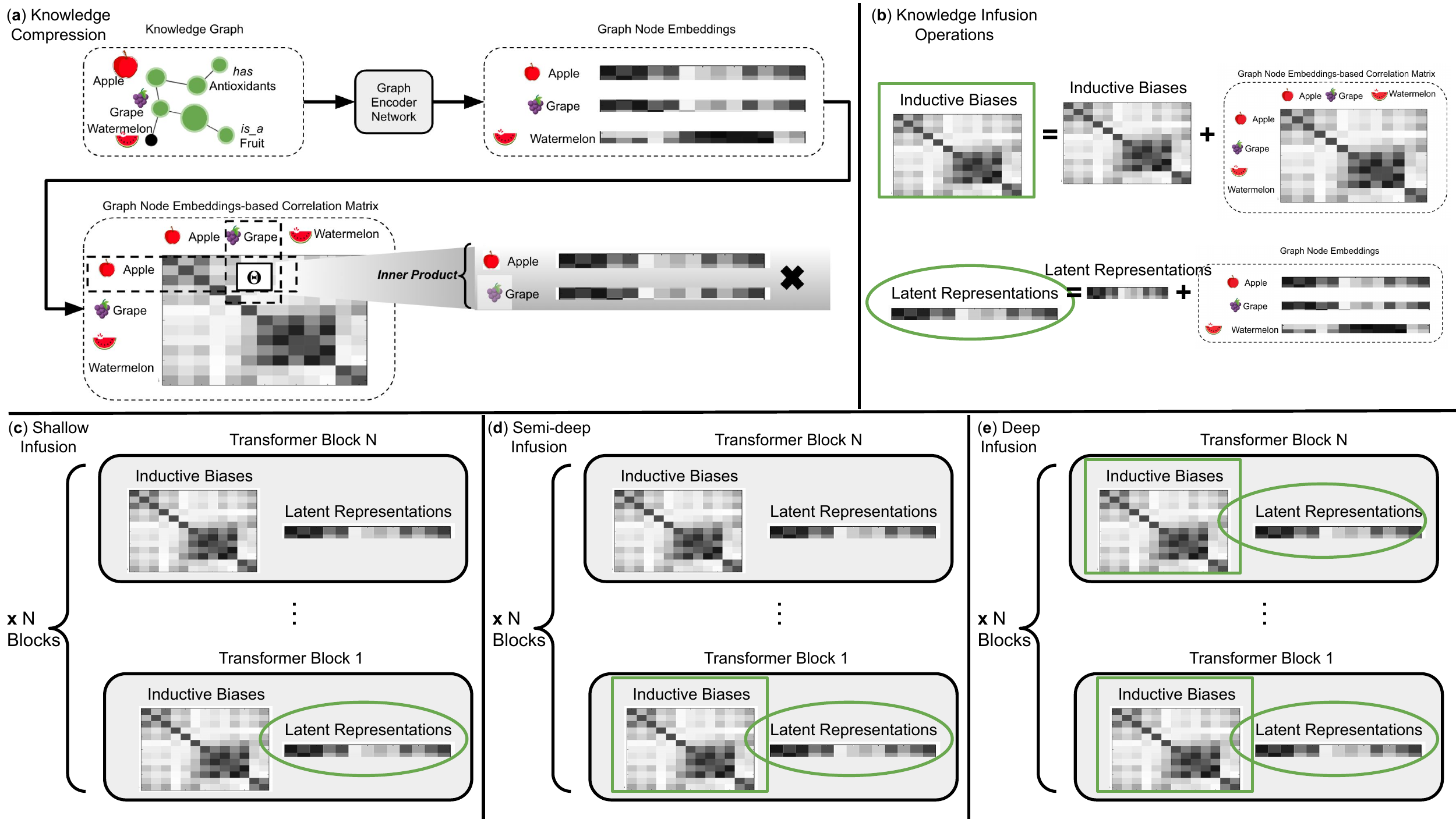}
    \caption{(a) Knowledge compression - Compressing the information in the knowledge graph into graph node embeddings (vectors) and graph node embedding-based correlations (matrices) for infusion in transformer architectures, namely at the latent representations of the transformer (vectors), or the inductive biases (self-attention matrices). (b) Knowledge infusion operations - Once compressed, we define the operation of knowledge infusion using vectors and matrices as summing the inductive biases (matrices) with the graph node embedding-based correlations and summing the latent representations (vectors) with the graph node embeddings. (c) Shallow infusion - Performing the knowledge infusion operation of adding the graph node embeddings to the latent representations of the first transformer block. (d) Semi-deep infusion - Performing the knowledge infusion operation of adding the graph node embedding-based correlation matrix to the inductive bias matrix of the first transformer block. (e) Deep infusion - Performing both adding the graph node embeddings and graph node embedding-based correlations to the latent representations and inductive biases across all $N$ transformer blocks.}
    \label{fig:KSAT}
\end{figure*}
\subsection*{(a) Knowledge Graph Compression for Knowledge Infusion}
Since a transformer architecture consists of two distinct types of components, namely (i) inductive biases represented as matrices and (ii) latent representations represented as vectors (as depicted in Figure \ref{fig:transformer}), the challenge lies in compressing external knowledge from knowledge graphs into one of these mathematical representations. This compression process enables the infusion of external knowledge into the transformer-based models. For this, we first obtain knowledge graph node embeddings (vectors) using a graph encoder network (e.g., numberbatch embeddings obtained by compression of the knowledge graph ConceptNet) and then compute pair-wise inner products between the node embeddings to obtain a graph node embedding-based correlation matrix\cite{speer2017conceptnet}. Figure \ref{fig:KSAT} (a) illustrates this knowledge compression process.
\subsection*{(b) Defining Knowledge Infusion Operations}
Having obtained the knowledge graph representation in the form of matrices and vectors, we proceed to define two distinct infusion operations. The first operation involves infusing knowledge at the latent representations of a transformer block, which is achieved by adding the graph node embedding vectors to the existing latent representations within the transformer block. The second operation pertains to infusing knowledge at the inductive biases of a transformer block. In this case, the infusion entails adding the graph node embedding-based correlation matrix to the inductive bias, which corresponds to the self-attention matrix of the transformer block. Figure \ref{fig:KSAT} (b) provides a visual depiction of these two operations.
\subsection*{(c) Shallow Knowledge Infusion}
Shallow Infusion involves the operation of infusing the knowledge at the latent representations of just the first transformer block of a transformer-based architecture. It is important to note that a single block usually consists of multiple heads, each associated with its own set of latent representations. In such cases, the knowledge infusion operation is performed on all the latent representations corresponding to the multiple heads. Figure \ref{fig:KSAT} (c) visually represents the shallow infusion approach.
 \subsection*{(d) Semi-deep Knowledge Infusion}
 Semi-deep Infusion involves both the operations of infusing the knowledge at the latent representations and the inductive biases of just the first transformer block. Similar to the shallow infusion method, if there are multiple heads present, the infusion operations are performed across all heads. Figure \ref{fig:KSAT} (d) provides a visual illustration of the semi-deep infusion approach.
 \subsection*{(e) Deep Knowledge Infusion} 
 Deep Infusion involves both operations of infusing the knowledge at the latent representations and the inductive biases of all the $N$ transformer blocks. Once again, if there are multiple heads present, these operations are performed across all heads. Figure \ref{fig:KSAT} (e) illustrates the deep-infusion approach, showcasing the infusion of knowledge throughout the transformer blocks.
 \section{Evaluation Tasks and Metrics Used in Experimentation}\label{sec:eval}
\subsection{Evaluation Tasks}
To assess the performance, we utilize the General Language Understanding Evaluation (GLUE) benchmark tasks. We categorize the tasks we experiment with into three distinct types, which serve as significant indicators of a model's language comprehension abilities.:
\paragraph{Natural Language Inference (NLI) tasks}The GLUE tasks MNLI (Multi-genre Natural Language Inference), QNLI (Question Answering Natural Language Inference), and WNLI (Winograd Natural Language Inference) test NLI capabilities from varying angles. MNLI tests whether the model can appropriately judge if a sentence logically follows from another, i.e., logical entailment. QNLI tests similar logical entailment between question and statement pairs - does it make logical sense to ask a follow-up question? WNLI tests logical entailment in the presence of pronouns and the nouns they reference.
\paragraph{Textual Entailment (TE) tasks} The GLUE task RTE (Recognizing Textual Entailment) tests for logical entailment similar to the NLI task MNLI. However, RTE emphasizes on the meaning - given two text fragments, whether the meaning of one can be entailed (or can be inferred) from the other.
\paragraph{Textual Similarity (TS) tasks}The GLUE task QQP (Quora Question Pairs) tests for the ability to assess the semantic equivalence, measured as the similarity between a pair of questions that appear on the social media forum Quora. 
\begin{table*}[!htb] 
\begin{tabular}{c|p{1.2cm}|p{1.2cm}|p{1.2cm}|p{1.2cm}|p{1.2cm}}
\toprule[1.5pt]

Model                    & QQP  & QNLI & WNLI & MNLI & RTE     \\ \hline 
Baseline (XLNET)     & 74.79          & 84.17          & 79.9          & 72.3   & 83.6 \\
\hline
Shallow Infusion    & 76.11          & 89.9          & 80.5          & 75.3 &  85.1 \\ 
\hline 
Semi-Deep Infusion   & 80.2 & 90.5 & 90.5  & 82.1 & 90.3\\ 
\hline 
Deep-Infusion   & \textbf{80.9}          & \textbf{92.3}          & \textbf{90.91}          & \textbf{88.53} &  \textbf{90.4 }\\ 
\bottomrule[1.5pt] 
\end{tabular}
\caption{Shows the comparison of accuracy across the different GLUE tasks from Section \ref{sec:eval} for the baseline XLNET model and its variants using the different kinds of knowledge infusion - Shallow, Semi-deep and Deep Infusion. We see that there is steady improvement as external knowledge is included at different components, i.e., latent representations and Inductive biases of a transformer-based model.}
\label{tab:quant_disc}
\end{table*}
\subsection{Evaluation Metrics}\label{subsec:eval_met}
 In Section \ref{sec:intro}, we discussed how ad hoc knowledge infusion techniques can lead to models exploiting statistical artifacts towards achieving high downstream task performance. Therefore, to evaluate the different knowledge infusion methods, in addition to the traditional performance metrics of accuracy, F1-scores across GLUE tasks, we also devise the following metrics:
\subsubsection{\textbf{Combined Graph Encoder and KSAT model Accuracy (CGKA)}}\label{subsubsec:cgka}:This metric evaluates the combined accuracies of two components. Firstly, it measures the accuracy of link prediction by the graph encoder on a separate test set that includes triples from the knowledge graph it encodes. Link prediction is carried out by summing the subject and predicate vectors and finding the closest object vector in a (subject, predicate, object) triple (we check if the similarity is greater than a threshold in our experiments, see Section \ref{subsec:new_metrics}). These triples were not used during runtime. Secondly, it calculates the average accuracy of the KSAT model across all the GLUE tasks. The underlying idea is that if we observe high scores on conventional accuracy and F1-score metrics but low scores on the CGKA metric, it suggests that the KSAT model has possibly learned misleading patterns. A low CGKA score indicates a lack of capturing knowledge graph information by the graph encoder.
 \subsubsection{\textbf{Data Efficiency at K (DE@k)}}: This metric evaluates the performance of the KSAT model after training it with only k\% of the total available training data. The rationale behind this is that effective knowledge infusion should result in improved data sufficiency. In other words, the additional context provided by high-quality knowledge should compensate for the need for a large volume of data to achieve good performance.
\section{Knowledge Graphs, Graph Encoder Networks, and Transformer Models Used in Experimentation}\label{sec:models}
We use the knowledge graphs ConceptNet and WorNet in our experiments. For the graph encoder network to obtain graph node embeddings (see Figure \ref{fig:KSAT} (a)), we use ConceptNet Numberbatch embeddings and ewise embeddings for ConceptNet and WordNet, respectively\cite{speer2017conceptnet,kumar2019zero}. We sum the graph node embeddings from both ConceptNet and WordNet for every input token to obtain a single graph node embedding per token. For the transformer models using which we test knowledge infusion, we use the language models - BERT, XLNET, RoBERTa, ELECTRA, and Longformer\cite{devlin2018bert,liu2019roberta,clark2020electra,yang2019xlnet,beltagy2020longformer}.
\section{Experiments}
\subsection{Traditional Performance Tests}
For all our experiments, we use the large version of the models. All experiments are run on a single A100 GPU. We use the standard configurations of the models (e.g., 12 heads per block and $N$=12 in BERT)
Out of all the transformer-based models we experiment with from among those listed in Section \ref{sec:models}, XLNET performs the best, both as a baseline and when used with various knowledge infusion techniques. Table \ref{tab:quant_disc} shows this result.
\subsection{Performance Tests using Newly Introduced Metrics}\label{subsec:new_metrics}
For the DE@K metric, we want to see if higher performance can be achieved using lesser data. So we plot results averaged across the GLUE tasks for each transformer-based model with the different infusion techniques using only 50\% of the training data, i.e., DE@50. We also want to measure the combined accuracies on link prediction of the embeddings from ConceptNet and WordNet(summed) with the average performance across the GLUE tasks. The method of link prediction is as described in Section \ref{subsubsec:cgka}, i.e., to predict a link between a subject and object in a (subject, predicate, object) triple, we check if the cosine similarity between the sum of subject and predicate vectors, and the object vectors is greater than 0.5 (we tuned this number from the set \{0, 0.25, 0.5, 0.75\}. Table \ref{tab:san} shows the results.
\begin{table}[!htb]
\footnotesize
\centering
\begin{tabular}{p{1cm}|p{1.25cm}|p{1.25cm}|p{1.25cm}|p{1.25cm}} 
\toprule[1.5pt]
Model                                                                                      & CGKA 
& DE@50 & DE@75 & DE@100 \\
\midrule[1pt] \\
BERT & 80/71/67/ &  81/79/70 & 81/81/73 & 82/81/73\\
\hline \\
RoBERTa & 79/67/65/ &  81/74/71 & 81/76/73 & 82/79/73\\
\hline \\
ELECTRA & 80/67/60/ & 75/72/72 & 77/73/71 & 81/75/71\\
\hline \\
XLNET & 81/71/65/ & 80/79/69 & 82/80/71 & 83/81/73\\
\hline \\
Longformer & 78/67/61/ & 79/75/73 & 79/76/72 & 78/76/74\\
\bottomrule[1.5pt]
\end{tabular}
\caption{Here, we see the performance (accuracy) of different models with deep-infusion/semi-deep infusion/shallow-infusion using the metrics introduced in Section \ref{sec:eval}. We see that both metrics improve with the addition of external knowledge using the infusion methods, with deep infusion performing the best. Furthermore, we see that the accuracy numbers are already in the 70-80s with just 50\% of the data points used (this is average accuracy measured across all GLUE tasks)}.
\label{tab:san}
\end{table}
\section{Conclusion and Future Work}
This paper introduces a systematic approach to knowledge infusion in transformer-based models. The findings indicate that incorporating external knowledge indeed enhances the performance of the models on language understanding tasks. Even more so as the infusion techniques operate on both the latent representations and inductive biases of the model across all transformer layers. This improvement is observed through the evaluation of both traditional metrics and newly introduced evaluation metrics, validating the effectiveness of knowledge infusion. For future research, we plan to explore hybrid knowledge infusion, which involves selectively choosing the blocks where knowledge infusion occurs and determining which knowledge to infuse. This differs from the current setup where graph node embeddings from WordNet and ConceptNet are summed. The analysis presented in this paper aims to lay the foundation for more principled approaches to external knowledge-augmented language models in the future. 
\section{Acknowledgement}

This work is built on prior work \cite{roy2023process,roy2023demo,roy2021depression,roy2021bknowledge,asawa2020covid,roy2022ksat,venkataramanan2023cook,roy2023knowledge,gaur2021can,roy2023proknow,tsakalidis2022overview,gupta2022learning,dolbir2021nlp,rawte2022tdlr,lokala2021edarktrends,zi2023ierl}, and supported by the National Science Foundation under Grant 2133842, “EAGER: Advancing Neuro-symbolic AI with Deep Knowledge-infused Learning" \cite{sheth2023neurosymbolic,sheth2021knowledge,sheth2022process}.
\bibliographystyle{unsrt}
\bibliography{references}
\end{document}